\documentclass{svproc}

\usepackage{amsmath} 
\usepackage{graphicx}
\usepackage{multirow}
\usepackage[caption=false]{subfig}
\usepackage[hidelinks]{hyperref}
\usepackage[dvipsnames]{xcolor}
\graphicspath{{./img/}}

\newcommand{\argmax}{\arg\!\max}
\newcommand\norm[1]{\left\lVert#1\right\rVert}

\begin{document}
	\mainmatter              
	\title{An Analysis of Word2Vec for the Italian Language}
	\titlerunning{An Analysis of Word2Vec for the Italian Language}  
	\author{	Giovanni Di Gennaro\inst{1}, Amedeo Buonanno\inst{2}, Antonio Di Girolamo\inst{1}, \\ 
			Armando Ospedale\inst{1}, Francesco A.N. Palmieri\inst{1}, Gianfranco Fedele\inst{3}}
	\authorrunning{Di Gennaro, Buonanno, Di Girolamo, Ospedale, Palmieri, Fedele} 
	\tocauthor{Di Gennaro, Buonanno, Di Girolamo, Ospedale, Palmieri, Fedele}
	\institute{	Universit\'a degli Studi della Campania ``Luigi Vanvitelli'', \\ Dipartimento di Ingegneria \\
	              	via Roma 29, Aversa (CE), Italy\\
	              	\and
	              	ENEA, Energy Technologies Department - Portici Research Centre,\\
		 	P. E. Fermi, 1, Portici (NA), Italy;\\
			\and
	              	MAZER s.r.l., Marcianise (CE), Italy;\\
	\email{ 	\{giovanni.digennaro, francesco.palmieri\}@unicampania.it \\
			amedeo.buonanno@enea.it\\
		     	\{antonio.digirolamo, armando.ospedale\}@studenti.unicampania.it\\
			gianfranco.fedele@laila.tech	}
	}
	\maketitle   

	\begin{abstract}
		Word representation is fundamental in NLP tasks, because it is precisely from the coding of semantic closeness between words that it is possible to think of teaching a machine to understand text. Despite the spread of word embedding concepts, still few are the achievements in linguistic contexts other than English. In this work, analysing the semantic capacity of the Word2Vec algorithm, an embedding for the Italian language is produced. Parameter setting such as the number of epochs, the size of the context window and the number of negatively backpropagated samples is explored.
		
	\end{abstract}

	\section{Introduction}
		In order to make human language comprehensible to a computer, it is obviously essential to provide some word encoding. The simplest approach is the one-hot encoding, where each word is represented by a sparse vector with dimension equal to the vocabulary size. In addition to the storage need, the main problem of this representation is that any concept of word similarity is completely ignored (each vector is orthogonal and equidistant from each other). On the contrary, the understanding of natural language cannot be separated from the semantic knowledge of words, which conditions a different closeness between them. Indeed, the semantic representation of words is the basic problem of Natural Language Processing (NLP). Therefore, there is a necessary need to code words in a space that is linked to their meaning, in order to facilitate  a machine in potential task of  ``understanding" it. In particular, starting from the seminal work \cite{Bengio2001}, words are usually  represented as dense distributed vectors that preserve their uniqueness but, at the same time, are able to encode the similarities. 

These word representations are called Word Embeddings since the words (points in a space of vocabulary size) are mapped in an embedding space of lower dimension. Supported by the distributional hypothesis \cite{Harris1954} \cite{Firth1957}, which states that a word can be semantically characterized based on its context (i.e. the words that surround it in the sentence), in recent years many word embedding representations have been proposed (a fairly complete and updated review can be found in \cite{Almeida2019} and \cite{Zhang2016}). These methods can be roughly categorized into two main classes: {\em prediction-based models} and {\em count-based models}. The former is generally linked to work on Neural Network Language Models (NNLM) and use a training algorithm that predicts the word given its local context, the latter leverage word-context statistics and co-occurrence counts in an entire corpus. The main prediction-based and count-based models are respectively Word2Vec \cite{Mikolov2013} (W2V) and GloVe \cite{GloVe}. 

Despite the widespread use of these concepts \cite{Schnabel2015} \cite{Bakarov2018}, few  contributions exist regarding the development of a W2V that is not in English. In particular, no detailed analysis on an Italian W2V seems to be present in the literature, except for \cite{AIWE} and \cite{WEGI}. However, both seem to leave out some elements of fundamental interest in the learning of the neural network, in particular relating to the number of epochs performed during learning, reducing the importance that it may have on the final result. In \cite{AIWE}, this for example leads to the simplistic conclusion that (being able to organize with more freedom in space) the more space is given to the vectors, the better the results may be. However, the problem in complex structures is that large embedding spaces can make training too difficult.

In this work, by setting the size of the embedding to a commonly used average value, various parameters are analysed as the number of learning epochs changes, depending on the window sizes and the negatively backpropagated samples.

	\section{Word2Vec}
\begin{figure}[h!]
			\centering
			\includegraphics[width=0.5\linewidth]{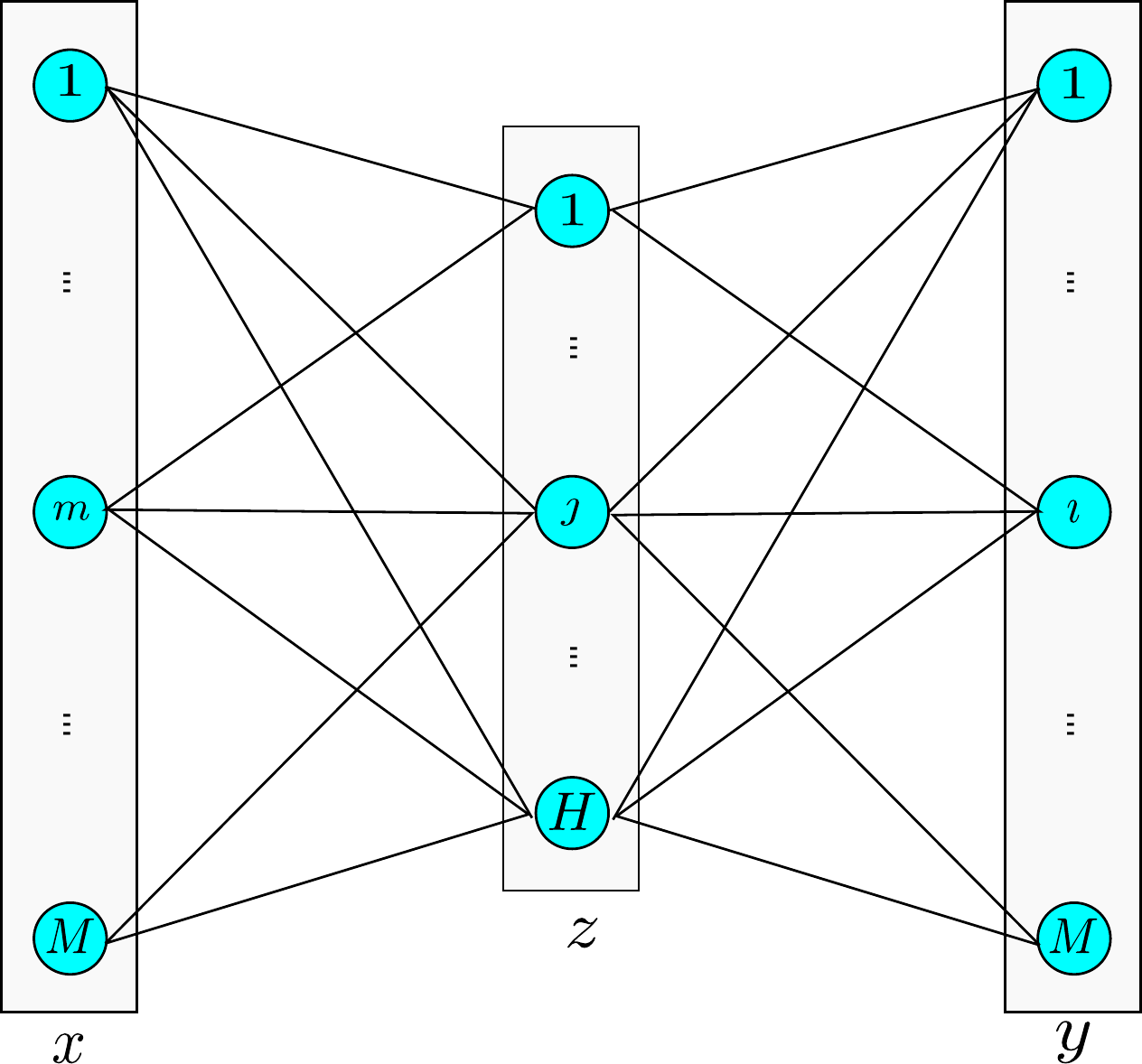}
			\caption{Representation of Word2Vec model.}
			\label{fig:w2vBase}
		\end{figure}

		The W2V structure consists of a simple two-level neural network (Figure \ref{fig:w2vBase}) with  one-hot vectors representing words at the input. It can be trained in two different modes, algorithmically similar, but different in concept: Continuous Bag-of-Words (CBOW) model and Skip-Gram model. While CBOW tries to predict the target words from the context, Skip-Gram instead aims to determine the context for a given target word. The two different approaches therefore modify only the way in which the inputs and outputs are to be managed, but in any case, the network does not change, and the training always takes place between single pairs of words (placed as one-hot in input and output).

The text is in fact divided into sentences, and for each word of a given sentence a window of words is taken from the right and from  the left to define the context. The central word is coupled with each of the words forming the set of pairs for training. Depending on the fact that the central word represents the output or the input in training pairs, the CBOW and Skip-gram models are obtained respectively.

Regardless of whether W2V is trained to predict the context or the target word, it is  used as a word embedding in a substantially different manner from the one for which it has been trained. In particular, the second matrix is totally discarded during use, since the only thing relevant to the representation is the space of the vectors generated in the intermediate level (embedding space).

		\subsection{Sampling rate}
			The common words (such as ``the", ``of", etc.) carry very little information on the target word with which they are coupled, and through backpropagation they tend to have extremely small representative vectors in the embedding space. To solve both these problems the W2V algorithm implements a particular ``subsampling" \cite{Mikolov2}, which acts by eliminating some words from certain sentences. Note that the elimination of a word directly from the text means that it no longer appears in the context of any of the words of the sentence and, at the same time, a number of pairs equal to (at most) twice the size of the window relating to the deleted word will also disappear from the training set. 

In practice, each word is associated with a sort of ``keeping probability" and, when you meet that word, if this value is greater than a randomly generated value then the word will not be discarded from the text. The W2V implementation assigns this ``probability" to the generic word $w_i$ through the formula:

			\begin{equation}
  				P(w_i) = \left(\sqrt{\frac{f(w_i)}{s}} + 1 \right) \frac{s}{f(w_i)}, \label{eq:sampling_rate}
			\end{equation}

\noindent
where $f(w_i)$ is the relative frequency of the word $w_i$ (namely $count(w_i)/total$), while $s$ is a sample value, typically set between $10^{-3}$ and $10^{-5}$.

		\subsection{Negative sampling}	
			Working with one-hot pairs of words means that the size of the network must be the same at input and output, and must be equal to the size of the vocabulary. So, although very simple, the network has a considerable number of parameters to train, which lead to an excessive computational cost if we are supposed to backpropagate all the elements of the one-hot vector in output. 

The ``negative sampling" technique \cite{Mikolov2} tries to solve this problem by modifying only a small percentage of the net weights every time. In practice, for each pair of words in the training set, the loss function is calculated only for the value $1$ and for a few values $0$ of the one-hot vector of the desired output. The computational cost is therefore reduced by choosing to backpropagate only $K$ words ``negative" and one positive, instead of the entire vocabulary. Typical values for negative sampling (the number of negative samples that will be backpropagated and to which therefore the only positive value will always be added), range from 2 to 20, depending on the size of the dataset.

The probability of selecting a negative word to backpropagate depends on its frequency, in particular through the formula:

			\begin{equation}
  				P(w_i) = \frac{f(w_i)^{3/4}}{\sum_{j=0}^n f(w_j)^{3/4}}
			\end{equation}

\noindent
Negative samples are then selected by choosing a sort of ``unigram distribution", so that the most frequent words are also the most often backpropated ones.

	\section{Implementation details}
		The dataset needed to train the W2V was obtained using the information extracted from a dump of the Italian Wikipedia (dated 2019.04.01), from the main categories of Italian Google News (WORLD, NATION, BUSINESS, TECHNOLOGY, ENTERTAINMENT, SPORTS, SCIENCE, HEALTH) and from some anonymized chats between users and a customer care chatbot (Laila\footnote[1]{\url{https://laila.tech/}}). The dataset (composed of 2.6 GB of raw text) includes $421\,829\,960$ words divided into $17\,305\,401$ sentences.

The text was previously preprocessed by removing the words whose absolute frequency was less than $5$ and eliminating all special characters. Since it is impossible to represent every imaginable numerical value, but not wanting to eliminate the concept of ``numerical representation" linked to certain words, it was also decided to replace every number present in the text with the particular $\langle NUM \rangle$ token; which probably also assumes a better representation in the embedding space (not separating into the various possible values). All the words were then transformed to lowercase (to avoid a double presence) finally producing a vocabulary of $618\,224$ words. 

\newpage
Note that among the special characters are also included punctuation marks, which therefore do not appear within the vocabulary. However, some of them (`.', `?' and `!') are later removed, as they are used to separate the sentences.

The Python implementation provided by Gensim was used for training the various embeddings all with  size 300 and sampling parameter ($s$ in Equation \ref{eq:sampling_rate}) set at $0.001$.

	\section{Results}
		To analyse the results we chose to use the test provided by \cite{WEGI}, which consists of $19\,791$ analogies divided into $19$ different categories: $6$ related to the ``semantic" macro-area ($8915$ analogies) and $13$ to the ``syntactic" one ($10876$ analogies). All the analogies are composed by two pairs of words that share a relation, schematized with  the equation: $a:a^{*}=b:b^{*}$ (e.g. ``man : woman = king : queen"); where $b^{*}$ is the word to be guessed (``queen"), $b$ is the word coupled to it (``king"), $a$ is the word for the components to be eliminated  (``man"), and $a^{*}$ is the word for the components to be added (``woman").

The determination of the correct response was obtained both through the classical additive cosine distance (3COSADD) \cite{Mikolov2013}:
		\begin{equation}
  			\argmax_{b^{*} \in V} \cos(b^{*}, b - a + a^{*})
		\end{equation}
\noindent
and through the multiplicative cosine distance (3COSMUL) \cite{levy2014}:

		\begin{equation}
  			\argmax_{b^{*} \in V} \frac{\cos(b^{*}, b)  \cos(b^{*}, a^{*})}{\cos(b^{*}, a) + \epsilon}
		\end{equation}
\noindent
where $\epsilon=10^{-6}$ and $\cos(x, y) = \frac{x \cdot y}{\norm{x}\norm{y}}$. The extremely low value chosen for the $\epsilon$ is due to the desire to minimize as much as possible its impact on performance, as during the various testing phases we noticed a strange bound that is still being investigated. As usual, moreover, the representative vectors of the embedding space are previously normalized for the execution of the various tests.

		\subsection{Analysis of the various models}
			We first analysed $6$ different implementations of the Skip-gram model each one trained  for $20$ epochs. Table \ref{tab:accuracy20} shows the accuracy values (only on possible analogies) at the 20th epoch for the six models both using 3COSADD and 3COSMUL. It is interesting to note that the 3COSADD total metric, respect to 3COSMUL, seems to have slightly better results in the two extreme cases of limited learning (W5N5 and W10N20) and under the semantic profile. However, we should keep in mind that the semantic profile is the one best captured by the network in both cases, which is probably due to the nature of the database (mainly composed of articles and news that principally use an impersonal language). In any case, the improvements that are obtained under the syntactic profile lead to the 3COSMUL metric obtaining better overall results. 

\begin{figure}[h!]
    				\centering
				\subfloat[]{{\includegraphics[width=.45\linewidth]{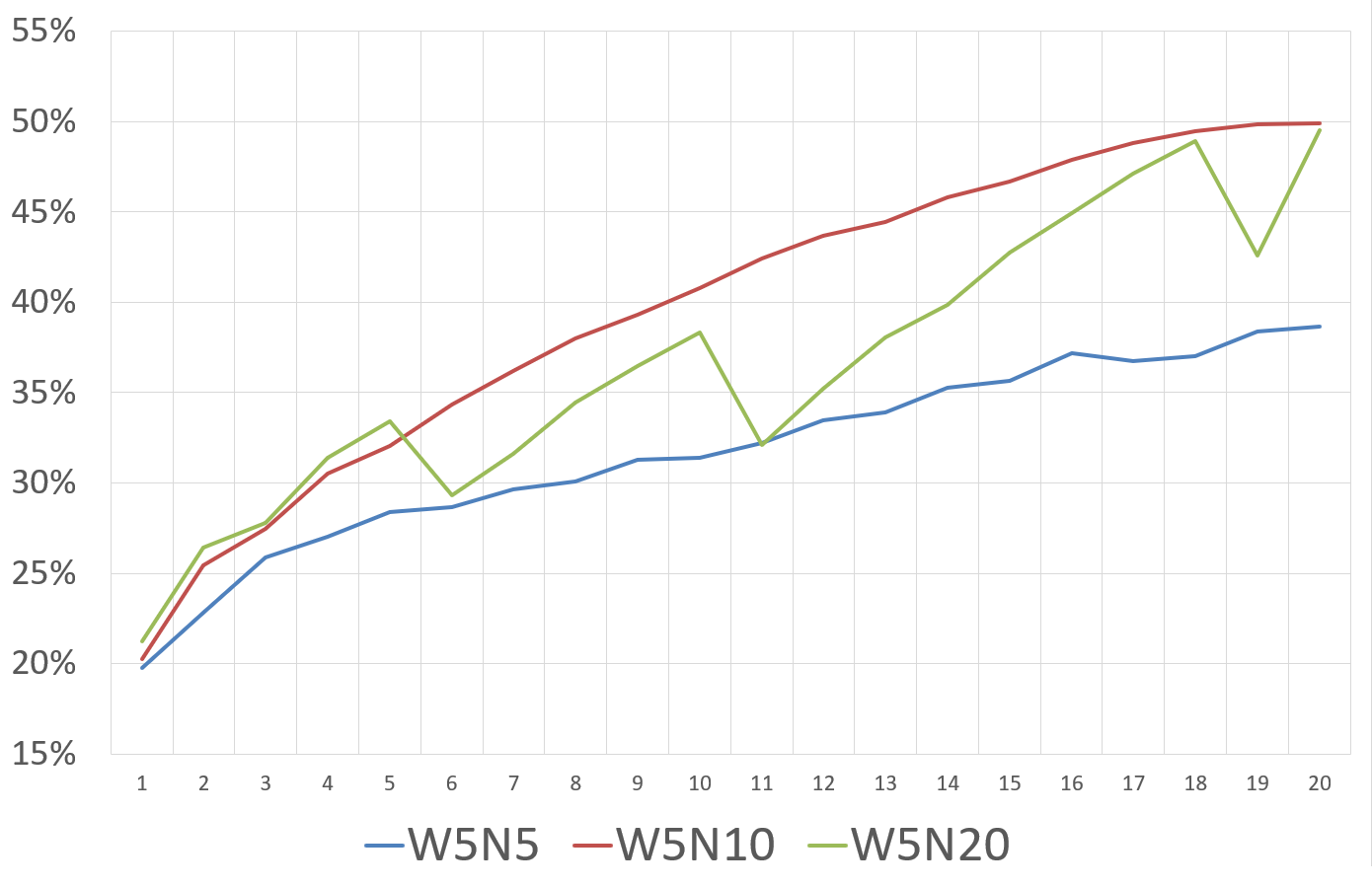} }}
				\qquad
				\subfloat[]{{\includegraphics[width=.45\linewidth]{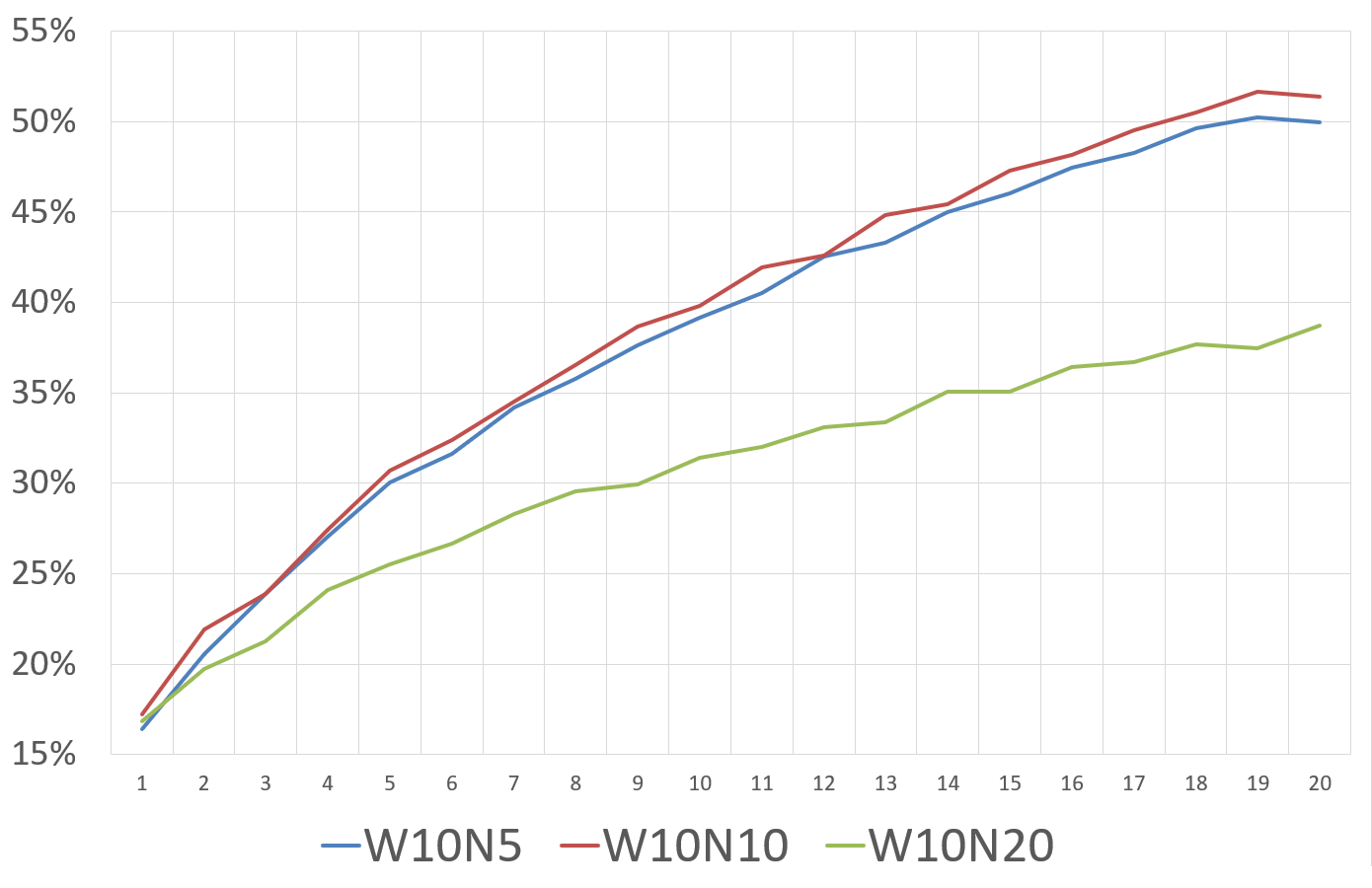} }}
				\vspace{-0.5em}
				\caption{Total accuracy using 3COSMUL at different epochs with negative sampling equal to 5, 10 and 20, where: (a) window is 5 and (b) window is 10.}
				\label{fig:testTotal20}
				\vspace{-1em}
			\end{figure}

Figure \ref{fig:testTotal20} shows the trends of the total accuracy at different epochs for the various models using 3COSMUL (the trend obtained with 3COSADD is very similar). Here we can see how the use of high negative sampling can worsen performance, even causing the network to oscillate (W5N20) in order to better adapt to all the data. The choice of the negative sampling to be used should therefore be strongly linked to the choice of the window size as well as to the number of training epochs.

			\begin{table}[h!]
				\vspace{-1em}
				\begin{center}
					\setlength\tabcolsep{0.5em}
					\begin{tabular}{l|l|c|c|c|c|c|c}
						\cline{3-8}
						\multicolumn{2}{ c }{} & \multicolumn{3}{ c| }{3COSADD} & 
						\multicolumn{3}{ c }{3COSMUL\rule{0pt}{12pt}} \\[2pt] \cline{3-8}
						\multicolumn{2}{ c }{} & Semantic & Syntactic & Total & Semantic & Syntactic &
						\multicolumn{1}{ c }{Total\rule{0pt}{12pt}} \\[2pt]
						\hline
						\multirow{3}{*}{W = 5} & N = 5 & {\bf 40,93\%} & {\bf 38,85\%} & {\bf 39,85\%} 
						& 39,62\% & 37,78\% & 38,67\%\rule{0pt}{12pt}\\
						& N = 10 & 52,99\% & 45,57\% & 49,14\% & {\bf 53,35\%} & {\bf 46,71\%} & {\bf 49,91\%}\\
						& N = 20 & {\bf 53,66\%} & 44,06\% & 48,68\% & 53,56\% & {\bf 45,80\%} & {\bf 49,53\%}\\[2pt]
						\hline
						\multirow{3}{*}{W = 10} & N = 5 & {\bf 53,63\%} & 45,55\% & 49,44\% 
						& 53,39\% & {\bf 46,79\%} & {\bf 49,97\%}\rule{0pt}{12pt}\\
						& N = 10 & \textcolor{red}{\bf 55,76\%} & 45,92\% & 50,66\% 
						& 55,56\% & \textcolor{red}{\bf 47,54\%} & \textcolor{red}{\bf 51,40\%}\\
						& N = 20 & {\bf 45,56\%} & {\bf 34,95\%} & {\bf 40,06\%} & 44,35\% & 33,52\% & 38,74\%\\[2pt]
						\hline
					\end{tabular}
				\end{center}
				\caption{Accuracy at the 20th epoch for the 6 Skip-gram models analysed when the W dimension of the window and the N value of negative sampling change.}
				\label{tab:accuracy20}
				\vspace{-4em}
			\end{table}

\begin{figure}[h!]
    				\centering
				\subfloat[]{{\includegraphics[width=.45\linewidth]{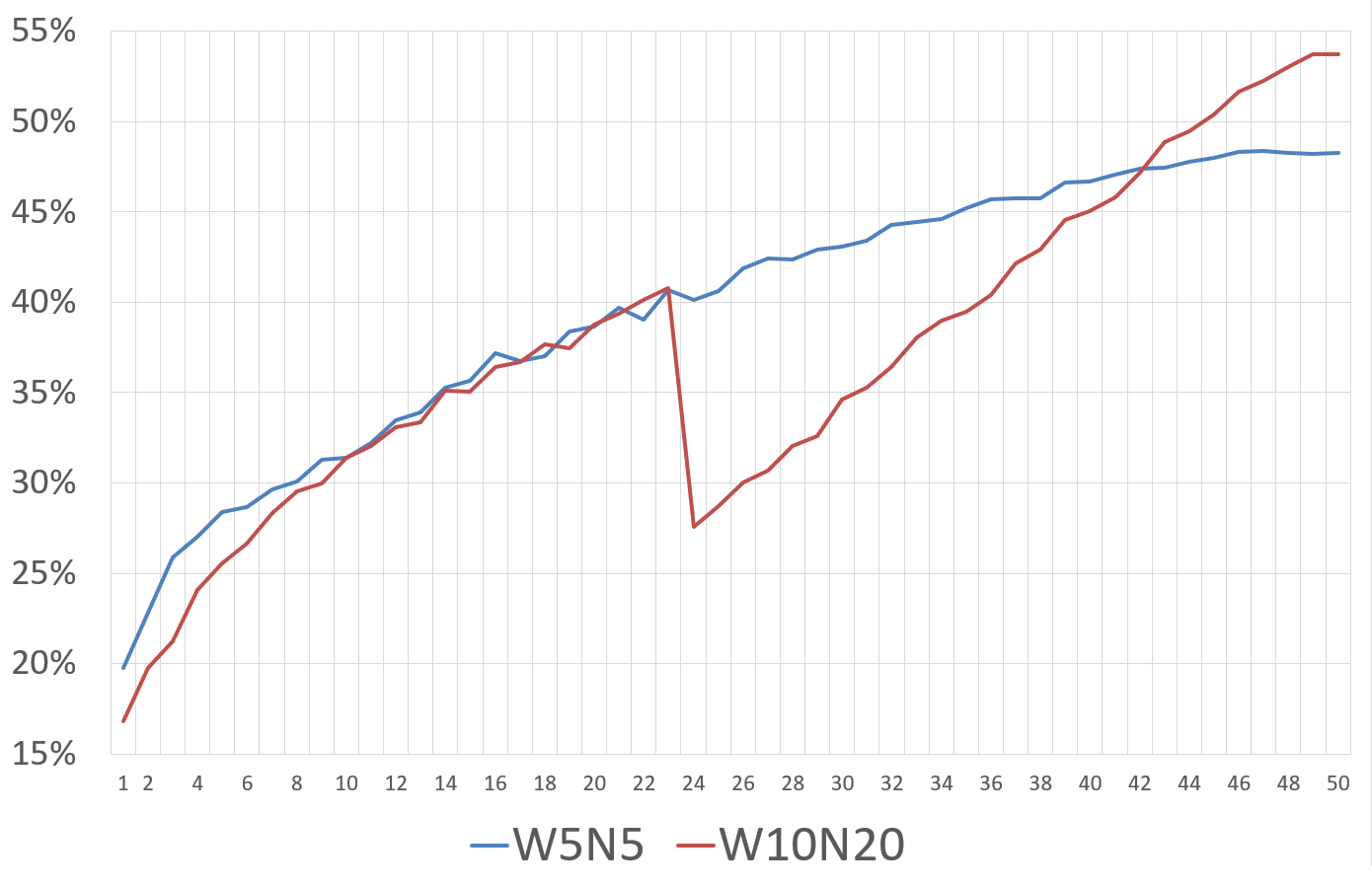} }}
				\qquad
				\subfloat[]{{\includegraphics[width=.45\linewidth]{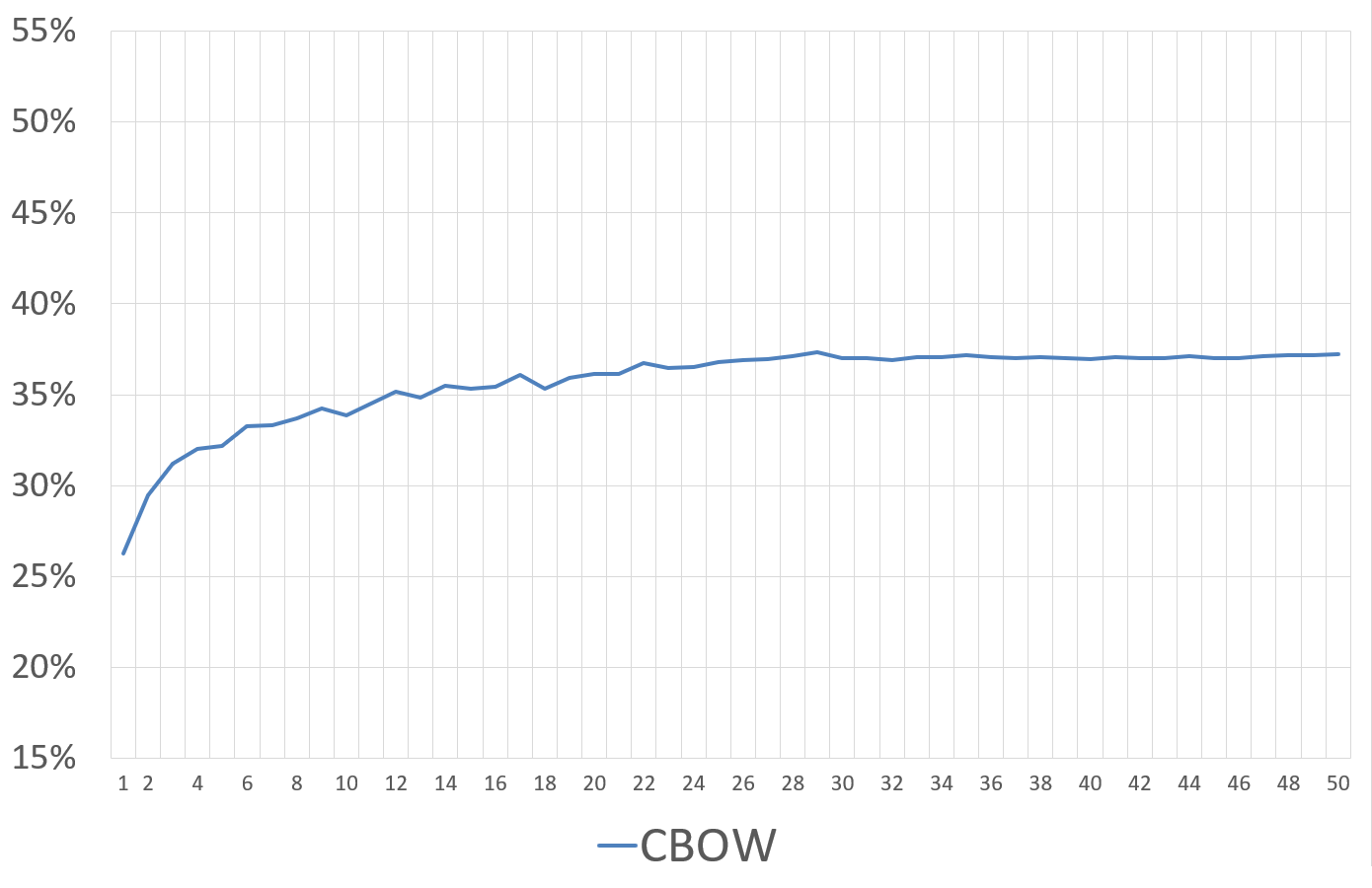} }}
				\vspace{-0.5em}
				\caption{Total accuracy using 3COSMUL up to the 50th epoch for: (a) the two worst Skip-gram models and (b) CBOW model with $W=10$ and $N=20$}
				\label{fig:testTotal50}
			\end{figure}

Continuing the training of the two worst models up to the 50th epoch, it is observed (Table \ref{tab:accuracy50}) that they are still able to reach the performances of the other models. The W10N20 model at the 50th epoch even proves to be better than all the other previous models, becoming the reference model for subsequent comparisons. As the various epochs change (Figure \ref{fig:testTotal50}.a) it appears to have the same oscillatory pattern observed previously, albeit with only one oscillation given the greater window size. This model is available at: \url{https://mlunicampania.gitlab.io/italian-word2vec/}.

			\begin{table}[h!]
				\begin{center}
					\setlength\tabcolsep{0.5em}
					\begin{tabular}{l|c|c|c|c|c|c}
						\cline{2-7}
						\multicolumn{1}{ c }{} & \multicolumn{3}{ c| }{3COSADD} & 
						\multicolumn{3}{ c }{3COSMUL\rule{0pt}{12pt}} \\[2pt] \cline{2-7}
						\multicolumn{1}{ c }{} & Semantic & Syntactic & Total & Semantic & Syntactic &
						\multicolumn{1}{ c }{Total\rule{0pt}{12pt}} \\[2pt]
						\hline
						W5N5 & 49,59\% & 45,25\% & 47,34\% 
						& {\bf 49,78\%} & {\bf 46,84\%} & {\bf 48,26\%}\rule{0pt}{12pt}\\[2pt]
						\hline
						W10N20 & \textcolor{red}{\bf 59,20\%} & 46,98\% & 52,86\% 
						& 59,07\% & \textcolor{red}{\bf 48,80\%} & \textcolor{red}{\bf 53,74\%}\rule{0pt}{12pt}\\[2pt]
						\hline
					\end{tabular}
				\end{center}
				\caption{Accuracy at the 50th epoch for the two worst Skip-gram models.}
				\label{tab:accuracy50}
				\vspace{-2em}
			\end{table}

\newpage
Various tests were also conducted on CBOW models, which however proved to be in general significantly lower than Skip-gram models. Figure \ref{fig:testTotal50}.b shows, for example, the accuracy trend for a CBOW model with a window equal to 10 and negative sampling equal to $20$, which on $50$ epochs reaches only $37.20\%$ of total accuracy (with 3COSMUL metric).

		\subsection{Comparison with other models}
			Finally, a comparison was made between the Skip-gram model W10N20 obtained at the 50th epoch and the other two W2V in Italian present in the literature (\cite{AIWE} and \cite{WEGI}). The first test (Table \ref{tab:totalCompare}) was performed considering all the analogies present, and therefore evaluating as an error any analogy that was not executable (as it related to one or more words absent from the vocabulary).

				\begin{table}[h!]
					\vspace{-1em}
					\begin{center}
						\setlength\tabcolsep{1em}
						\begin{tabular}{c|c|c|c|c}
							\cline{3-5}
							\multicolumn{2}{ c }{} & Semantic & Syntactic & 
							\multicolumn{1}{ c }{Total\rule{0pt}{12pt}} \\[2pt]
							\hline
							\multirow{3}{*}{3COSADD} &  {\bf Our model} &  {\bf 58,42\%} &  {\bf 40,92\%} &  {\bf 48,81\%}\rule{0pt}{12pt}\\
							& Tipodi’s model \cite{AIWE} & 53,21\% & 37,37\% & 44,51\%\\
							& Berardi's model \cite{WEGI} & 48,81\% & 32,62\% & 39,91\%\\[2pt]
							\hline
							\multirow{3}{*}{3COSMUL} &  {\bf Our model} &  {\bf 58,31\%} &  {\bf 42,51\%} &  {\bf 49,62\%} \rule{0pt}{12pt}\\
							& Tipodi’s model \cite{AIWE} & 55,56\% & 39,60\% & 46,79\%\\
							& Berardi's model \cite{WEGI} & 49,59\% & 33,70\% & 40,86\%\\[2pt]
							\hline
						\end{tabular}
					\end{center}
					\caption{Accuracy evaluated on the total of all the analogies}
					\label{tab:totalCompare}
					\vspace{-2em}
				\end{table}

As it can be seen, regardless of the metric used, our model has significantly better results than the other two models, both overall and within the two macro-areas. Furthermore, the other two models seem to be more subject to the metric used, perhaps due to a stabilization not yet reached for the few training epochs.

For a complete comparison, both models were also tested considering only the subset of the analogies in common with our model (i.e. eliminating from the test all those analogies that were not executable by one or the other model). Tables \ref{tab:tipodiCompare} and \ref{tab:berardiCompare} again highlight the marked increase in performance of our model compared to both.

				\begin{table}[!]
					\vspace{-1em}
					\begin{center}
						\setlength\tabcolsep{1em}
						\begin{tabular}{c|c|c|c|c}
							\cline{3-5}
							\multicolumn{2}{ c }{} & Semantic & Syntactic & 
							\multicolumn{1}{ c }{Total\rule{0pt}{12pt}} \\[2pt]
							\hline
							\multirow{2}{*}{3COSADD} &  {\bf Our model} &  {\bf 59,20\%} &  {\bf 47,95\%} &  {\bf 53,43\%} \rule{0pt}{12pt}\\
							& Tipodi’s model \cite{AIWE} & 53,92\% & 43,94\% & 48,81\%\\[2pt]
							\hline
							\multirow{2}{*}{3COSMUL} &  {\bf Our model} &  {\bf 59,08\%} &  {\bf 49,75\%} &  {\bf 54,30\%}\rule{0pt}{12pt}\\
							& Tipodi’s model \cite{AIWE} & 56,30\% & 46,57\% & 51,31\%\\[2pt]
							\hline
						\end{tabular}
					\end{center}
					\caption{Accuracy evaluated only on the analogies common to both vocabularies}
					\label{tab:tipodiCompare}
					\vspace{-2.5em}
				\end{table}

				\begin{table}[!]
					\vspace{-1em}
					\begin{center}
						\setlength\tabcolsep{1em}
						\begin{tabular}{c|c|c|c|c}
							\cline{3-5}
							\multicolumn{2}{ c }{} & Semantic & Syntactic & 
							\multicolumn{1}{ c }{Total\rule{0pt}{12pt}} \\[2pt]
							\hline
							\multirow{2}{*}{3COSADD} & {\bf Our model} &  {\bf 59,20\%} &  {\bf 48,48\%} &  {\bf 53,73\%} \rule{0pt}{12pt}\\
							& Berardi's model \cite{WEGI} & 49,45\% & 38,73\% & 43,98\%\\[2pt]
							\hline
							\multirow{2}{*}{3COSMUL} & {\bf Our model} &  {\bf 59,08\%} &  {\bf 50,35\%} &  {\bf 54,63\%} \rule{0pt}{12pt}\\
							& Berardi's model \cite{WEGI} & 50,25\% & 40,00\% & 45,02\%\\[2pt]
							\hline
						\end{tabular}
					\end{center}
					\caption{Accuracy evaluated only on the analogies common to both vocabularies}
					\label{tab:berardiCompare}
					\vspace{-2.5em}
				\end{table}

	\section{Conclusion}
		In this work we have analysed the Word2Vec model for Italian Language obtaining a substantial increase in performance respect to other two models in the literature (and despite the fixed size of the embedding). These results, in addition to the number of learning epochs, are probably also due to the different phase of data pre-processing, very carefully excuted  in performing a complete cleaning of the text and above all in substituting the numerical values with a single particular token.
		We have observed that the number of epochs is an important parameter and its increase leads to  results that rank our two worst models almost equal, or even better than others. 

		Changing the number of epochs, in some configurations, creates an oscillatory trend, which seems to be linked to a particular interaction between the window size and the negative sampling value.
		In the future, thanks to the collaboration in the Laila project, we intend to expand the dataset by adding more user chats. The objective will be  to verify if the use of a less formal language can improves accuracy in the syntactic macro-area.
	
	\newpage
	\bibliographystyle{ieeetr}
	\bibliography{paperBib}

\end{document}